\documentclass[10pt,twocolumn,letterpaper]{article}

\usepackage[pagenumbers]{cvpr} 

\usepackage{amsmath,amssymb,amsfonts}
\usepackage{algorithmic}
\usepackage{textcomp}
\usepackage{booktabs}
\usepackage{pifont}
\usepackage{multirow}
\usepackage{fontawesome}
\usepackage{bbding}
\usepackage{scalefnt}
\usepackage{graphicx}
\usepackage[hyphens]{url}
\usepackage{bm}

\usepackage{xcolor}
\definecolor{cvprblue}{rgb}{0.21,0.49,0.74}
\usepackage[pagebackref,breaklinks,colorlinks,citecolor=cvprblue]{hyperref}

\def\paperID{*****} 
\def\confName{CVPR}
\def\confYear{2024}

\title{WeNLEX: Weakly Supervised Natural Language Explanations for Multilabel Chest X-ray Classification}

\author{Isabel Rio-Torto\thanks{Correspondence to: \texttt{isabel.riotorto@inesctec.pt}}  \textsuperscript{1,2}, Jaime S. Cardoso\textsuperscript{1,2}, Luís F. Teixeira\textsuperscript{1,2}
\\
\small\textsuperscript{1}INESC TEC
\quad\textsuperscript{2}Universidade do Porto
}

\begin{document}
\maketitle

\begin{abstract}
Natural language explanations provide an inherently human-understandable way to explain black-box models, closely reflecting how radiologists convey their diagnoses in textual reports. Most works explicitly supervise the explanation generation process using datasets annotated with explanations. Thus, though plausible, the generated explanations are not faithful to the model’s reasoning. In this work, we propose \emph{WeNLEX}, a weakly supervised model for the generation of natural language explanations for multilabel chest X-ray classification. Faithfulness is ensured by matching images generated from their corresponding natural language explanations with original images, in the black-box model’s feature space. Plausibility is maintained via distribution alignment with a small database of clinician-annotated explanations. We empirically demonstrate, through extensive validation on multiple metrics to assess faithfulness, simulatability, diversity, and plausibility, that \emph{WeNLEX} is able to produce faithful and plausible explanations, using as little as 5 ground-truth explanations per diagnosis. Furthermore, \emph{WeNLEX} can operate in both \textit{post-hoc} and in-model settings. In the latter, i.e., when the multilabel classifier is trained together with the rest of the network, \emph{WeNLEX} improves the classification AUC of the standalone classifier by 2.21\%, thus showing that adding interpretability to the training process can actually increase the downstream task performance. Additionally, simply by changing the database, \emph{WeNLEX} explanations are adaptable to any target audience, and we showcase this flexibility by training a layman version of \emph{WeNLEX}, where explanations are simplified for non-medical users.
\end{abstract}

\section{Introduction}
\label{sec:introduction}
The importance of explainability, particularly in critical domains like medicine~\cite{Vayena,rudin2019stop} has motivated research into diverse explanation modalities, such as Natural Language Explanations (NLEs). NLEs are textual descriptions that go beyond image captioning, since besides the input image they also need to take into account the underlying model’s decision-making process~\cite{ibpria,faithvsplaus,zsnles}. Unlike visual explanations, which are spatially precise but often semantically limited, NLEs are inherently human-understandable, and they are the explanation modality of choice in certain medical contexts where clinicians (e.g., radiologists) convey their findings through textual reports~\cite{ibpria,miller,gale}.

Despite their promise, current NLE generation approaches face critical limitations. Most rely on fully human-annotated explanation datasets with one NLE per image per diagnosis. This not only imposes a significant annotation burden but also assumes that the reasoning of the model being explained (MBE) is similar to the reasoning of the human annotators. In other words, fully supervised NLEs reflect the annotations, i.e., are plausible, but not the model’s true reasoning, i.e., are not faithful. This reliance on explicit supervision also prevents fully explaining pretrained models, i.e., produce \textit{post-hoc} explanations, as it would require ground-truth NLEs for every combination of correct/incorrect predictions, which differs from model to model. In fact, being able to explain incorrect model decisions is crucial, e.g., for debugging purposes, and it constitutes a clear case where the model reasoning does not align with the human's, hence the definition of an incorrect model prediction.

To address these challenges, we propose \emph{WeNLEX} (pronounced ``weenlex''), the first weakly supervised framework for generating NLEs for multilabel chest X-ray classification. \emph{WeNLEX} ensures faithfulness by matching the images generated from the NLEs with the original input images, in the MBE’s feature space. Plausibility is maintained through distribution alignment with a small set of clinician-annotated explanations. Through exhaustive validation across multiple dimensions (faithfulness, simulatability, diversity, and
plausibility), we show that \emph{WeNLEX}: 
\begin{enumerate}
    \item generates faithful NLEs, requiring as few as five ground-truth explanations per diagnosis
    \item can operate in both \textit{post-hoc} and in-model settings, even improving classification performance by 2.21\% AUC, demonstrating that interpretability can enhance task performance
    \item offers adaptability to different target audiences, exemplified by a layman version that produces simplified explanations for non-expert users
\end{enumerate}

\section{Related Work}
\label{sec:related_work}

The literature on the generation of NLEs for image-only or image-text tasks is vast~\cite{pjx, fme, esnlive, rvt, evil, rexc, ofax, nlxgpt}. Hence, there are many architectures for NLE generation, which can be grouped according to different criteria:
\begin{itemize}
    \item network architecture (the type of neural network): convolutional neural networks~\cite{pjx,fme,esnlive,gift,wojciechowski} vs. vision transformers~\cite{rvt,evil,rexc,ofax,nlxgpt,gift,uninlx} for image processing, and recurrent neural networks~\cite{pjx,fme} vs. transformer-based models~\cite{rvt,evil,rexc,gift,wojciechowski,uninlx} for text generation
    \item modularity (separation between task prediction and explanation modules): modular~\cite{pjx,fme,rvt,evil,rexc,gift,wojciechowski} vs. unified~\cite{nlxgpt,ofax,uninlx}
    \item training paradigm (MBE trained separately or jointly with the explanation model): \textit{post-hoc}~\cite{pjx,fme,rvt,evil,gift,wojciechowski} vs. in-model~\cite{rexc,nlxgpt,ofax,uninlx}
    \item reasoning flow (order between prediction and explanation): \emph{predict-explain}~\cite{pjx,fme,rvt,evil,nlxgpt,ofax,uninlx,gift,wojciechowski} vs. \emph{explain-predict}~\cite{rexc}
\end{itemize}

In the medical domain there are, to the best of our knowledge, only three works on NLE generation~\cite{oana,kgllava,midl}. Kayser \textit{et al.}~\cite{oana} introduce the MIMIC-NLE dataset and test different architectures to generate NLEs for multilabel chest X-ray classification, all with the same working principles: i) a vision model extracts
features and classifies the input image, and ii) for each predicted diagnosis, the features, the multilabel prediction vector, and the pathology for which the NLE is being generated, are given to a language model to autoregressively generate the NLE.

More recently, Hamza \textit{et al.}~\cite{kgllava} proposed a Knowledge Graph Retrieval Augmented Generation (KG-RAG) framework to enhance the LoRA-based~\cite{lora} finetuning of LLaVA~\cite{llava15}. They build a knowledge graph of relationships between medical entities extracted from the medical reports of MIMIC-CXR~\cite{mimic} using the RadGraph model~\cite{radgraph}. For each image, the closest KG entities are retrieved and given to the language module of LLaVa~\cite{llava15}, together with the projected image embeddings and the vision classifier predictions. By incorporating domain-specific knowledge, this approach achieves state-of-the-art results on MIMIC-NLE.

All aforementioned works fully supervise the NLE generation process, optimising cross-entropy between generated and human-annotated NLEs. As argued, this leads to plausible but unfaithful explanations that reflect annotations rather than the model’s reasoning. Sammani and Deligiannis~\cite{zsnles} are the first to tackle this problem in a zero-shot manner. They train a multilayer perceptron (MLP) to map the space of a textual encoder into the vision classifier space. At inference time, learnable prefixes steer an off-the-shelf language model to generate NLEs that maximise the similarity with visual features through the text encoder+MLP. Regardless of being a plug-and-play approach, easily adaptable to any vision classifier, it might not be suitable for problems with less classes (originally the MLP is trained for 1000 classes and it is not clear if it would converge for a smaller number of classes), and it cannot operate in multilabel settings.

Rio-Torto \textit{et al.}~\cite{midl} propose replacing the commonly used Decoder-only NLE generator with an Encoder-Decoder architecture and showed that it is possible to supervise NLE generation directly in the Encoder latent space. This allows imposing desirable properties on the NLEs via the continuous Encoder latent space, thus avoiding Reinforcement Learning (RL) when teacher forcing is not possible (e.g., when no, or few, ground-truth NLEs are available). In this work, we extend~\cite{midl} to the weakly supervised setting, thus avoiding the pitfalls of fully supervising the generation of NLEs.

\section{Methodology}
\label{sec:method}

\subsection{MIMIC-NLE Dataset}
\label{subsec:dataset}
In the general domain, image captioning datasets abound~\cite{coco,sharma2018cc3m,changpinyo2021cc12m}. Datasets with NLEs are more scarce, but do exist, both for text-only~\cite{esnli,cage,comve,ecqa} and vision-language (VL) tasks~\cite{pjx,esnlive,evil}. Wiegreffe and Marasovi{\'c}~\cite{datasets} provide a comprehensive review on the topic. In the medical domain the same trend emerges: there are some datasets that include medical reports~\cite{mimic,chexpertplus,padchest,bimcv}, but datasets with NLEs are rarer. In fact, MIMIC-NLE~\cite{oana} is, as far as we are aware, the only dataset with NLEs for chest X-ray classification.

The MIMIC-NLE dataset is automatically extracted from MIMIC-CXR~\cite{mimic} reports using clinically validated rules and the CheXbert labeler~\cite{chexbert}. It comprises 38003 image-NLE pairs or 44935 image-label-NLE triplets (one NLE can explain multiple diagnoses/labels). Each image can have up to 10 ($L = 10$) pathologies simultaneously, each with 3 possibilities ($C = 3$): \textit{Positive} (clear evidence of the presence of the pathology), \textit{Uncertain} (the pathology might be present), and \textit{Negative} (clear evidence of the absence of that pathology). These include: i) diagnoses labels, i.e., the labels being explained by the NLEs (\textit{Atelectasis}, \textit{Consolidation}, \textit{Edema}, \textit{Pleural Effusion}, \textit{Pleural Other}, \textit{Pneumonia}, and \textit{Pneumothorax}), and ii) evidence labels, i.e., labels that are part of the evidence that some diagnosis label is present in the input image (\textit{Consolidation}, \textit{Enlarged Cardiomediastinum}, \textit{Lung Lesion}, and \textit{Lung Opacity}). \textit{Consolidation} can be considered both a diagnosis or an evidence label, depending on the interaction with other pathologies (e.g., consolidation as a diagnosis label or consolidation as evidence for the existence of pneumonia).

A key limitation of this dataset is the strong imbalance in evidence labels, particularly \textit{Lung Opacity} (around 77\% of NLEs). Additionally, because evidence extraction relies on the CheXbert labeler~\cite{chexbert}, around 15\% of the NLEs do not have any evidence label; this may be because there are actually no evidence keywords in the original NLE (see examples A and B below) or simply that the evidence present in the NLE is not among the 14 labels that CheXbert is able to predict (see examples C and D below). In such cases evaluating the evidence in generated NLEs is not possible: even incorrect or uninformative outputs (examples A and B) could be deemed valid if CheXbert predicts no evidence labels. To address this, we remove instances without evidence labels. All experiments use this version of the dataset, comprising 36501 NLEs (35600 train, 254 val, 647 test) and 28548 images (27848 train, 198 val, 502 test).

\begin{center}
    \fcolorbox{black}{gray!20}{%
      \parbox{0.95\linewidth}{%
      \scalefont{0.85}
        A. Findings most consistent with moderate pulmonary edema.\\
        B. Findings suggesting mild pulmonary vascular congestion.\\
        C. There is again a coarse reticular abnormality favoring the bases and peripheral aspects of the lung, most consistent with pulmonary fibrosis.\\
        D. There is mild vascular congestion consistent with fluid overload.
        }%
    }
\end{center}

\begin{figure*}[!t]
\centering
\includegraphics[width=0.95\textwidth]{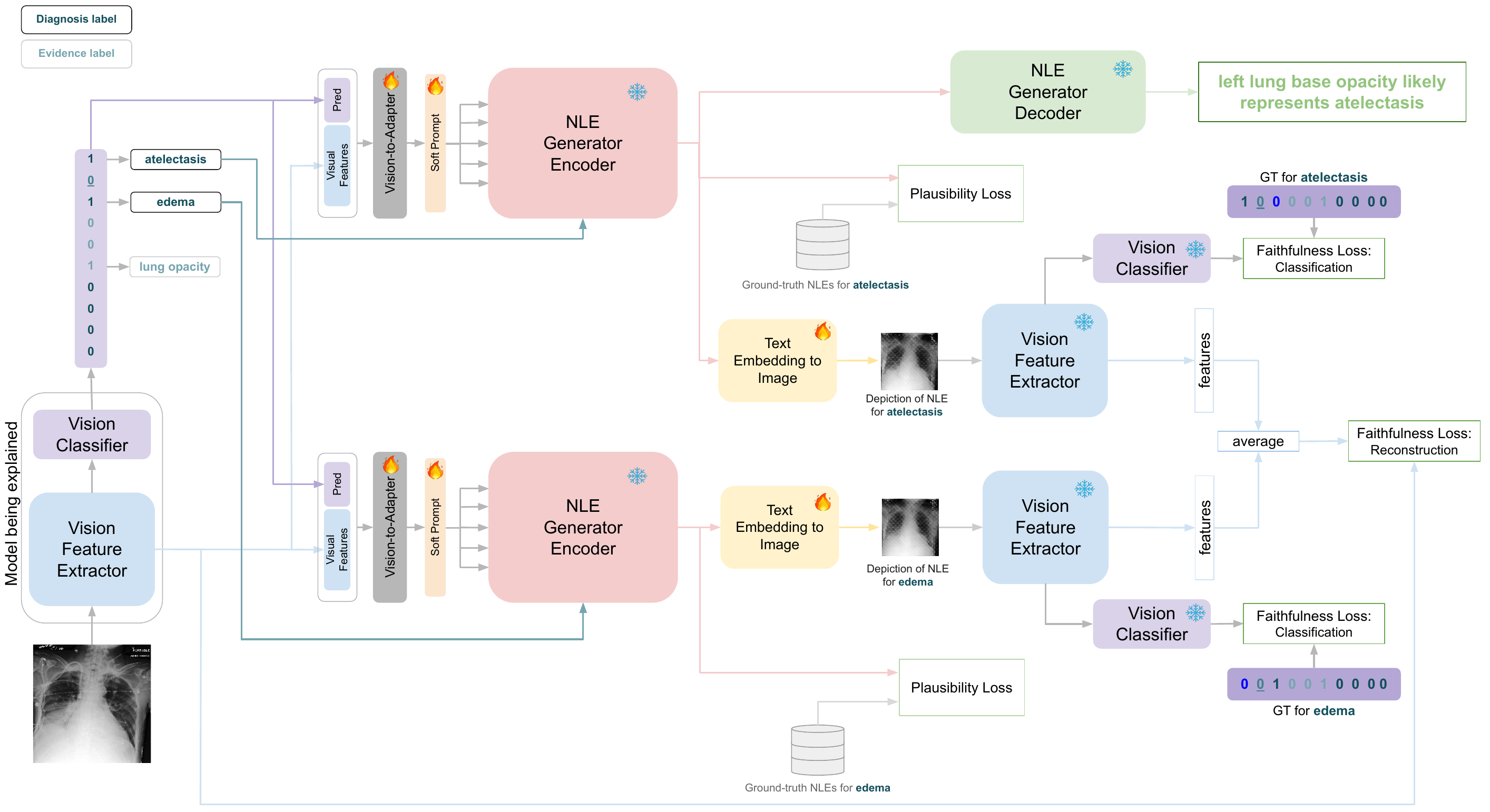}
\caption{Architecture of \emph{WeNLEX}, a weakly supervised model that generates natural language explanations (NLEs) for a multilabel X-ray classifier. For each predicted diagnosis (e.g., atelectasis, edema), it produces an NLE (only the atelectasis NLE is shown). A pretrained, frozen text-only Encoder–Decoder is adapted with soft prompt tuning (\emph{Vision-to-Adapter} and \emph{Soft Prompt}) to take as input the image features, the entire prediction vector (including diagnosis and evidence labels), and the textual label of the diagnosis being explained. The \emph{NLE Generator Encoder} outputs an NLE embedding, which is compared to ground-truth NLE embeddings for that diagnosis to ensure plausibility (\emph{Plausibility Loss}). Each NLE is also given to a \emph{Text Embedding to Image} model, which generates an image depicting its content. This image is then processed by the model being explained (MBE) to extract features. To enforce faithfulness, the average of these features across all NLEs for an image is compared with the original image features (\emph{Faithfulness Loss: Reconstruction}). Finally, each NLE must recover the MBE’s original diagnosis prediction: the MBE’s output for the generated image/NLE is compared against the original prediction (\emph{Faithfulness Loss: Classification}). Trainable layers/parameters are represented by the fire icon, while frozen blocks are represented by the snowflake.}
\label{fig:architecture}
\end{figure*}

\subsection{Problem Formulation}
Let us consider a multiclass multilabel classification task, where the goal is to predict a target $y \in \{1,...,C\}^L$, where $L$ is the number of labels (of which more than one can be present simultaneously) and $C$ is the number of classes per label, from an input image $x \in \mathbb{R}^{c \times h \times w}$, where $c, h, w$ are the number of channels, height and width, respectively. Let us also consider a classification model, $f_{\theta_1}$, that learns to predict $y$ from $x$, i.e., $\hat{y} = f_{\theta_1}(x)$. Our goal is to design a system, \emph{WeNLEX}, that generates explanations in natural language, $\hat{e}$, for the predictions of classifier $f$, such that, $\hat{e} = g_{\theta_2}(x, \hat{y}, f_{\theta_1})$. We adopt the predict-explain paradigm, since we generate one NLE per predicted label and, thus, need to know what the prediction is beforehand.

During training of \emph{WeNLEX} two scenarios are possible: i) $f_{\theta_1}$ has been trained and remains frozen (\textit{post-hoc} scenario), or ii) $f_{\theta_1}$ is being trained simultaneously with $g_{\theta_2}$ (in-model scenario). As will be further detailed in the following sections, the flexibility of \emph{WeNLEX} allows for both.

\subsection{WeNLEX}

Fig.~\ref{fig:architecture} presents the overall architecture of \emph{WeNLEX}, where for each predicted diagnosis label (either as uncertain or positive), an NLE is generated. \emph{WeNLEX} is inspired by work on unsupervised image captioning~\cite{unsupcapt} and it is built upon our previous work~\cite{midl}, where a pretrained text-only Transformer Encoder-Decoder model is adapted via Parameter-Efficient Fine-Tuning (PEFT) to receive the output of a multilabel classifier, its features, and the textual description of the diagnosis being explained. In this previous work, under the fully supervised setting (i.e., we had access to one ground-truth NLE per instance), we concluded that using the PEFT method called Multi-Modal LLaMA-Adapter~\cite{llama_adapter} achieved the best results, and that it was possible to supervise NLE generation at the sentence level (i.e., in the latent space of the Encoder), instead of at the word level (i.e., at the output of the Decoder). By doing this, one can impose desirable properties on the generated NLEs through the continuous Encoder latent space, thus avoiding the need for RL. Moreover, training is sped up because word-by-word decoding of the generated sentence embeddings into text is only done during inference. In this work, \emph{WeNLEX}, we extend this previous proposal to the weakly supervised scenario, not only to lower annotation costs, but also to mitigate the previously mentioned lack of faithfulness that arises from fully supervised NLE generation. Thus, \emph{WeNLEX} is designed considering the properties an NLE should have~\cite{gve,ibpria,jacovi,faithvsplaus}:
\begin{itemize}
    \item plausibility: sound coherent and logical to a human being
    \item faithfulness: reflect the model's decision process
    \item image-relevance: be specific to a given image
    \item adaptability: adapt to different users
\end{itemize}

In the following subsections we describe the proposed loss functions that promote each of the aforementioned properties.

\subsubsection{Plausibility}
In fully supervised NLE generation architectures, plausibility is achieved by approximating generated NLEs with ground-truth NLEs via, e.g., the cross entropy loss. In our case, to be able to do this in a weakly supervised manner (i.e., without a one-to-one correspondence between generated and ground-truth NLEs), we build a database with a fixed number of ground-truth NLEs per diagnosis label. In practice, this database does not have the ground-truth NLEs in textual form, but their embeddings given by the pretrained and now frozen NLE Generator Encoder. Therefore, this becomes a distribution matching problem between the generated and ground-truth NLEs. We experiment with two ways to tackle this: adversarial learning and Maximum Mean Discrepancy (MMD) minimization.

In the first approach, we use a Wasserstein Generative Adversarial Network with Gradient Penalty (WGAN-GP)~\cite{wgan, wgangp}. WGANP-GP has been proposed to mitigate the mode collapse problems of traditional GANs and is composed of: 
\begin{enumerate}
    \item the generator, $g_{\theta_2}$, generates an NLE for a predicted diagnosis conditioned on the visual features and the prediction of the vision classifier being explained ($f_{\theta_1}$). It learns to minimise the Wasserstein distance between the generated and real NLE embedding distributions by producing NLEs that the critic will score highly as belonging to the real data distribution.
    \item the discriminator (or critic), $d_{\theta_3}$, receives the generated NLE embedding and learns to estimate the Wasserstein distance between the real and generated NLE embedding distributions.
\end{enumerate}

Just like in the traditional GAN paradigm, the generator and discriminator play a min-max game, given by:
\begin{equation}
\min_{g} \max_{d \in \mathcal{D}_{1\text{-Lip}}} 
    \; \mathbb{E}_{e \sim \mathbb{P}_r}[d(e)] 
    - \mathbb{E}_{\hat{e} \sim \mathbb{P}_g}[d(\hat{e})]
\end{equation}

\noindent where $g$ is the NLE generator, $e$ is the ground-truth NLE embedding, $\hat{e}$ is the generated NLE embedding, $d$ is the discriminator, $\mathcal{D}_{1\text{-Lip}}$ is the set of all discriminators that are 1-Lipschitz continuous, $\mathbb{P}_r$ is the distribution of real NLE embeddings, and $\mathbb{P}_g$ is the distribution of generated NLE embeddings.

The generator loss is given by:
\begin{equation}
\mathcal{L}_g = - \mathbb{E}_{\hat{e} \sim \mathbb{P}_g}[d(\hat{e})]
\end{equation}

The discriminator loss is given by:
\begin{multline}
\mathcal{L}_d = \mathbb{E}_{\hat{e} \sim \mathbb{P}_g}[d(\hat{e})] 
     - \mathbb{E}_{e \sim \mathbb{P}_r}[d(e)] \\
     + \lambda \, \mathbb{E}_{\tilde{e} \sim \mathbb{P}_{\tilde{e}}} 
       \Big[ \big( \|\nabla_{\tilde{e}} d(\tilde{e})\|_2 - 1 \big)^2 \Big]
\end{multline}

\begin{equation*}
\tilde{e} = \alpha e + (1 - \alpha)\hat{e}, 
\quad \alpha \sim \mathcal{U}[0,1]
\end{equation*}

\noindent where $\tilde{e}$ is the interpolation between the ground-truth ($e$) and the generated ($\hat{e}$) NLE embeddings, $\mathbb{P}_{\tilde{e}}$ denotes the distribution of samples obtained by interpolation, and $\alpha$ is an interpolation factor sampled from a uniform distribution $\mathcal{U}$ on the interval $[0,1]$. $\nabla_{\tilde{e}} d(\tilde{e})$ is the gradient of the critic used to enforce the 1-Lipschitz constraint. This gradient penalty is controlled by the $\lambda$ hyperparameter.

Although WGANP-GP works well in theory, in practice it might be difficult to stabilize its training. Moreover, it introduces additional training parameters due to the discriminator. Therefore, we also experiment with MMD minimization.

MMD measures the distance between two distributions through the distance between their embeddings in the reproducing kernel Hilbert space (RKHS):

\begin{multline}
\widehat{\mathrm{MMD}}^2(X, Y) 
= 
\\
= \mathbb{E}_{i,j}[k(x_i, x_j)] 
+ \mathbb{E}_{i,j}[k(y_i, y_j)] 
- 2 \, \mathbb{E}_{i,j}[k(x_i, y_j)] \\
= \frac{1}{m^2} \sum_{i=1}^m \sum_{j=1}^m k(x_i, x_j)
+ \frac{1}{n^2} \sum_{i=1}^n \sum_{j=1}^n k(y_i, y_j) \\
- \frac{2}{mn} \sum_{i=1}^m \sum_{j=1}^n k(x_i, y_j)
\end{multline}

\noindent where $X$ is the generated NLE embeddings distribution, $Y$ is the ground-truth NLE embeddings distribution, $x_i$ and $x_j$ ($y_i$ and $y_j$) are samples from distribution $X$ ($Y$), $m$ ($n$) is the number of samples in $X$ ($Y$), and $k$ is a Gaussian kernel defined by $k(x, x') = \exp\!\left(-\frac{\|x - x'\|^2}{2\sigma^2}\right)$. The kernel computes all pairwise distances between $X$ and $Y$.

In practice, we compute MMD per-label and then average over all labels in the batch so, for each MMD computation, $X$ is composed of the NLEs generated for a given label, while $Y$ consists of the ground-truth NLEs for that label present in the NLE database.

\subsubsection{Faithfulness}
In addition to being plausible, NLEs should provide enough detail to allow the reconstruction of their corresponding images. Following~\cite{unsupcapt}, we do not attempt to reconstruct images in pixel space but instead in the feature space of the MBE. To achieve this, each generated NLE is passed to a module that converts its text embeddings into images. The latter are given to the MBE and the L2 distance between the original and the reconstructed image features is computed. Given the multilabel nature of our primary classification task, there can be several NLEs per image, each referring to a specific pathology. However, the features of the original image contain information on all predicted pathologies (i.e., the features are not disentangled by pathology). So, we actually measure the distance between original image features and the average of all features obtained from each generated image, i.e., from each generated NLE for a given input image (c.f. Fig~\ref{fig:architecture}).

Additionally, for an NLE to be faithful it also needs to be able to produce the same decision that gave rise to it in the first place. Although this is somewhat implicitly covered with the feature reconstruction loss detailed above, to ensure further disentanglement of the NLEs of a given input image, we also compute the classification loss for each generated image. However, two things are important to mention regarding the target of this classification loss: i) it is not the ground-truth of the original image, but the prediction vector of the classifier being explained (as previously explained the NLEs need to be faithful to the model, not to the human annotations), and ii) it is not the full prediction vector of the classifier, but only the prediction vector for that specific diagnosis (the evidence labels are the same for all targets, since there is no way of distinguishing which evidence label led to each diagnosis prediction).

Finally, in the in-model scenario, we ensure that the original classifier is not updated directly when it is used to compute the feature reconstruction and classification losses, i.e., it is updated but only through the backpropagation path through the Text Embedding to Image model and the NLE Generator Encoder. To achieve this, we maintain a frozen copy of the classifier to be used just for the feature extraction needed for the reconstruction and classification losses. Since the original classifier is being updated at every iteration and that could cause training instability, we only update the copy with the newest parameters every 1000 steps.

\subsubsection{Image-relevance} Image-relevance is guaranteed in two ways: by giving the visual features extracted by the MBE to the NLE Generator Encoder, and through the feature reconstruction loss, as explained above.

\subsubsection{Overall Loss Function}
\emph{WeNLEX} is trained with a combination of the three aforementioned loss functions: i) $\mathcal{L}_{nle\_plaus}$, the plausibility loss (either adversarial or MMD), ii) $\mathcal{L}_{nle\_recons}$, the image feature reconstruction loss, and iii) $\mathcal{L}_{nle\_clf}$, the NLE classification loss. In the in-model case, it also includes $\mathcal{L}_{img\_clf}$, the multiclass multilabel image classification loss, since the classifier is being trained alongside the NLE generation (i.e., in this case \emph{WeNLEX} can be considered a self-explanatory model).

The overall loss function uses the automatic loss weighting method of Cipolla \textit{et al.}~\cite{lossweights}, thus resulting in the following equations:

\begin{multline}
    \mathcal{L} = \frac{1}{2\sigma_1^2}\mathcal{L}_{plaus} + \frac{1}{2\sigma_2^2}\mathcal{L}_{nle\_clf}  + \frac{1}{2\sigma_3^2}\mathcal{L}_{nle\_recons} \\
    + log(1+\sigma_1^2) + log(1+\sigma_2^2) + log(1+\sigma_3^2)
\end{multline}

For the in-model case, the overall loss function includes the term $\frac{1}{2\sigma_4^2}\mathcal{L}_{img\_clf} + log(1+\sigma_4^2)$. The $\sigma_i$ are learnable parameters with an initial value of 1.

\section{Experiments}
\label{sec:experiments}

\subsection{Implementation Details}
Unless otherwise stated, all models are trained with the AdamW optimizer
for $50$ epochs with a batch size of $16$ and a linearly decayed learning rate of $5 \times 10^{-4}$ with $1000$ warmup steps. The final model is chosen based on the lowest validation loss.

\subsubsection{Multilabel Classifier}
Following previous work on the MIMIC-NLE dataset~\cite{oana,midl}, the vision classifier being explained is the DenseNet-121~\cite{densenet}. In the \textit{post-hoc} experiments, the classifier is pretrained on the images of the MIMIC-NLE dataset with a class-weighted cross entropy loss, obtaining an AUC of $65.13$.

\subsubsection{Text AutoEncoder}
The text autoencoder (NLE Generator Encoder and NLE Generator Decoder blocks of Fig.~\ref{fig:architecture}) is the same as in Rio-Torto \textit{et al}.~\cite{midl}, with the only difference being related to the pretraining dataset. In~\cite{midl} the encoder was the CXR-BERT model~\cite{cxrbert}, which is a masked language BERT-based transformer model trained on PubMed abstracts~\cite{pubmed}, clinical notes from MIMIC-III~\cite{mimiciii} and MIMIC-CXR~\cite{mimic}. Since in this work we are operating under the weakly supervised setting, we assume we do not have access to all MIMIC-NLE sentences. Considering that MIMIC-NLE has been derived from MIMIC-CXR, we wanted to make sure our encoder never had access to MIMIC-CXR. Thus, we take the PubMedBERT\footnote{\url{https://huggingface.co/microsoft/BiomedNLP-BiomedBERT-base-uncased-abstract}} (BERT model trained from scratch on PubMed abstracts only)~\cite{pubmedbert} and finetune it for 500k steps and an initial learning rate of $2\times10^{-5}$ on sentences of the Interpret-CXR dataset~\cite{interpretcxr} (from which we exclude MIMIC-CXR). Afterwards, we follow the procedure described in~\cite{midl,autobot} and train, still on Interpret-CXR, the single-layer transformer decoder with a denoising auto-encoder objective for 1M steps, batch size of $64$, and an initial learning rate of $1 \times 10^{-3}$. It achieves a reconstruction BLEU-4 score of $85.9$.

\subsubsection{Discriminator}
When using adversarial learning as the plausibility loss, the whole architecture of \emph{WeNLEX}, except the discriminator, is considered the generator (Vision-to-Adapter, NLE Generator Encoder, Text Embedding to Image model, etc). The discriminator is implemented as a lightweight feed-forward network that receives the NLE and a diagnosis-specific embedding (the latter is obtained by passing each diagnosis label through the frozen NLE Generator Encoder). The two 768-dimensional vectors are concatenated and passed through a series of fully connected layers with progressively reduced dimensionality ($2\times768$, $512$, $256$, and $1$). The hidden layers are followed by LeakyReLU activations ($\alpha = 0.2$). As is commonly done when training WGANs, for every 5 discriminator updates, the generator is updated once.

\subsubsection{Text Embedding to Image Model}
The Text Embedding to Image model converts the 768-dimensional NLE embeddings produced by the NLE Generator Encoder to $224 \times 224$ grayscale images. The embedding is first projected into a $512 \times 14 \times 14$ feature map using a fully connected layer with ReLU activation. It is then upsampled through four transposed convolutional layers, each doubling spatial resolution while reducing channels ($512 \rightarrow 256 \rightarrow 128 \rightarrow 64 \rightarrow 1$). Batch normalization and ReLU are applied after each intermediate layer, and a final Tanh activation normalizes the output image to $[-1,1]$.

\subsubsection{NLE Database}
To build the ground-truth NLE database, we randomly sample $n$ NLEs per diagnosis label, excluding NLEs that explain more than one diagnosis (e.g., ``New opacification at the right lung base could be either atelectasis or developing pneumonia.''), in order to obtain NLEs as specific to each diagnosis as possible.

\subsection{Evaluation Metrics}
Evaluating text generation tasks, e.g., image captioning, is notoriously difficult given the inherent variability and ambivalence of natural language, and still somewhat of an unsolved problem in the field~\cite{esnli,evil}. Evaluating NLEs is even harder because, e.g., compared to image captioning, another variable needs to be taken into account: the MBE.

\subsubsection{Plausibility}
Most works on NLE generation only measure the plausibility of NLEs by comparing them to human-annotated NLEs natural language generation metrics (BLEU~\cite{papineni2002bleu}, METEOR~\cite{meteor}, BERTScore~\cite{bertscore}, etc). Since BERTScore correlates higher with human evaluation~\cite{evil} and since our NLEs are from the medical domain, we compute their plausibility via BERTScore with the CheXbert~\cite{chexbert} model. This score (\textbf{CXBS}) is computed only for correctly predicted NLEs (correct diagnosis and evidence labels).

\subsubsection{Simulatability}
Another way of evaluating NLEs (or other kinds of explanations) is through their utility to an end-user, via simulatability: how well explanations help an observer reproduce the MBE's prediction~\cite{beenkim2017,las}.  This concept appears in the literature under different names and evaluation procedures~\cite{rexc,nlxgpt,wiegreffe21}. In NLX-GPT~\cite{nlxgpt} it is called ``Explain-Predict'' and the input question and the explanation are given to a language model, which is asked to answer the question (in the context of a visual question-answering task). Wiegreffe \textit{et al.}~\cite{wiegreffe21}, in the context of a text-only task, measure the additional ability to predict a label that an explanation provides over the input, i.e., the difference between task performance when an explanation is given together with the input vs. when it is not.
We adapt these approaches and present the $\bm{\hat{y}|NLE}$ and $\bm{\hat{y}|(img,NLE)}$ metrics, which measure whether the NLE alone or together with the image allow a human to reach the same prediction as the MBE. Following previous work~\cite{las}, we use a proxy for the human observer; we ask the VL foundation model CheXagent~\cite{chexagent} to answer if a given label is present in the NLE or in both the NLE and the image. We also compute $\bm{\hat{y}|img}$ to establish a baseline on the performance of CheXagent on the images alone.

\begin{table*}[ht]
\centering
\caption{Comparison of \emph{WeNLEX} with the fully supervised baseline of~\cite{midl} in terms of Faithfulness, Simulatability, Diversity, and Plausibility. \faFire \space refers to the in-model scenario - the MBE is trained alongside the rest of \emph{WeNLEX}), while \SnowflakeChevron \space refers to the \textit{post-hoc} scenario (MBE has been pretrained and is frozen during training of \emph{WeNLEX}). ``Clf.'' refers either to the image or the NLE classification loss. ``Plaus.'' refers to the plausibility loss and ``Recons.'' to the feature reconstruction loss. ``\# NLEs'' is the total number of generated NLEs; values in parentheses are for correctly classified cases (diagnosis and evidence). All metrics use the total \# NLEs, except CXBS, which uses only the parenthesised values.
The \textdownarrow \space means that for the diversity metrics lower is better. Best results are in bold; second-best are underlined.}
\label{tab:results}
\resizebox{\textwidth}{!}{%
\begin{tabular}{lccccccccccccccc} 
\toprule
\multirow{4}[7]{*}{Model} & \multicolumn{3}{c}{Image-Related} & \multicolumn{11}{c}{NLE-Related} \\
\cmidrule(lr){2-4}\cmidrule(lr){5-16}
& \multirow{3}[5]{*}{Clf.} & \multirow{3}[5]{*}{AUC} & \multirow{3}[5]{*}{$\hat{y}|img$} & \multirow{3}[5]{*}{Plaus.} & \multirow{3}[5]{*}{Recons.} & \multirow{3}[5]{*}{Clf.} & \multirow{3}[5]{*}{\# NLEs} & \multicolumn{3}{c}{Faithfulness} & \multicolumn{2}{c}{Simulatability} & \multicolumn{2}{c}{Diversity \textdownarrow} & Plausibility \\
\cmidrule(lr){9-11}\cmidrule(lr){12-13}\cmidrule(lr){14-15}\cmidrule{16-16}
& & & & & & & & & \multicolumn{2}{c}{Deletion} & \multirow{2}[3]{*}{$\hat{y}|NLE$} & \multirow{2}[3]{*}{$\hat{y}|(img,NLE)$} & \multirow{2}[3]{*}{S-BLEU} & \multirow{2}[3]{*}{Retrv.} & \multirow{2}[3]{*}{CXBS} \\
\cmidrule(lr){10-11}
& & & & & & & & CLEV (F1) & Flip (\%) & $\Delta p$\\ 
\midrule
\multicolumn{16}{c}{Fully Supervised} \\
\multicolumn{16}{c}{} \\
1 & \faFire & 64.08 & \textbf{83.29} & MSE & \ding{55} & \ding{55} & 1107 (128) & \underline{10.37} & \textbf{97.63} & \textbf{0.565} & 89.97 & 88.26  & 8.785 & 0.7263 & \underline{48.00} \\
\multicolumn{16}{c}{} \\ 
\midrule
\multicolumn{16}{c}{Weakly Supervised} \\
\multicolumn{16}{c}{} \\
2 & \multirow{4}{*}{\SnowflakeChevron} & \multirow{4}{*}{\underline{65.13}} & \multirow{4}{*}{\underline{82.47}} & Adv. & \ding{55} & \ding{55} & \multirow{4}{*}{1038 (128)} & 4.876 & 94.72 & 0.2671 & 71.77 & \underline{89.11} & 41.58 & 0.8079 & 36.21 \\
3 & & & & \multirow{4}{*}{MMD} & \ding{55} & \ding{55} & & 7.601 & 91.95 & 0.3192 & 86.89 & \textbf{89.21} & 10.36 & 0.6525 & \textbf{50.39} \\
4 & & & & & \ding{51} & \ding{55} & & 9.303 & 92.48 & 0.3236 & 77.84 & 86.71 & \textbf{2.671} & \textbf{0.5258} & 47.50 \\
5 & & & & & \ding{51} & \ding{51} & & 9.401 & 91.48 & \underline{0.3328} & \underline{90.46} & 88.82  & 4.037 & \underline{0.6286} & 47.35 \\
6 & \faFire & \textbf{67.34} & 79.91 & & \ding{51} & \ding{51} & 1115 (168) & \textbf{10.61} & \underline{95.65} & 0.3197 & \textbf{91.21} & 86.91 & \textbf{1.920} & 0.6704 & 45.38\\
\bottomrule
\end{tabular}
}
\end{table*}
\subsubsection{Faithfulness}
\begin{figure}[!tbh]
\centering
\includegraphics[width=0.98\columnwidth]{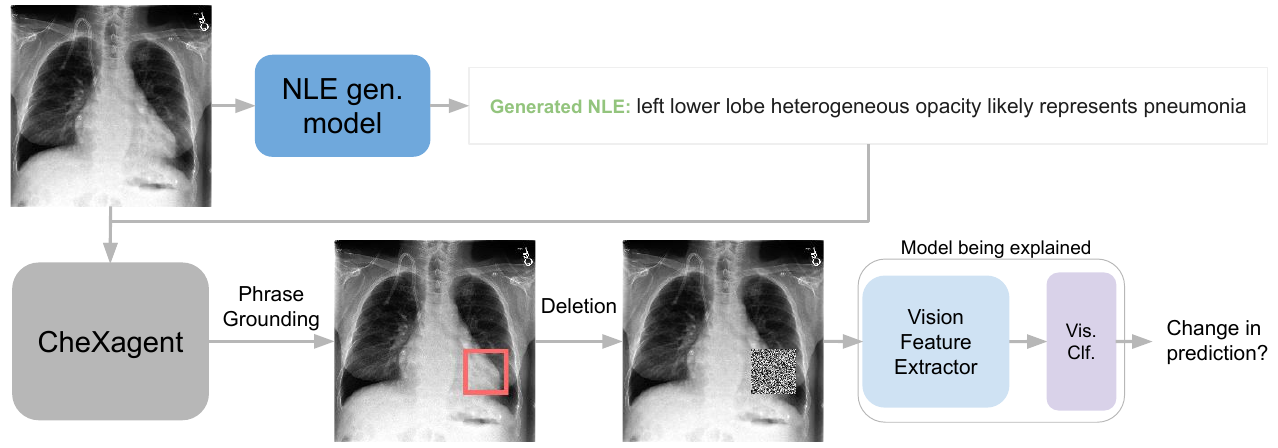}
\caption{Depiction of the deletion faithfulness metric: an image and a generated NLE are given to CheXagent, which grounds the text in the image. The identified regions are occluded, and the masked image is given to the model being explained (MBE). If the NLE is faithful, occluding these regions should significantly alter the MBE’s prediction.}
\label{fig:deletion}
\end{figure}

Most faithfulness metrics target setting with text-only inputs~\cite{oanafaith1,oanafaith2,faithvsplaus}. Both in the realm of visual explanations and NLEs for tasks where the input is text, a popular way to measure faithfulness is by removing parts of the input that the explanation considers important and measuring the decrease in the performance of the MBE~\cite{PetsiukDS18,deletion,oanafaith2}. Wojciechowski \textit{et al.}~\cite{wojciechowski} adapt this paradigm for the VL domain by asking humans to cover parts of the input image containing the decision rationale indicated in the NLE. We do the same but by leveraging the  
phrase grounding capabilities of CheXagent: we ask it to ground the NLE in the image and then mask the image in the location(s) given by the bounding boxe(s) produced by CheXagent (see Fig.~\ref{fig:deletion}). Following Wojciechowski \textit{et al.}~\cite{wojciechowski}, we measure the number of times the multilabel classifier flips its decision when given the masked image (\textbf{Flip (\%)}) and the absolute difference in the logits of the diagnosis originally predicted ($\bm{\Delta_p}$), considering both \textit{Uncertain} and \textit{Positive} logits together.

\begin{figure}[!h]
\centering
\includegraphics[width=0.98\columnwidth]{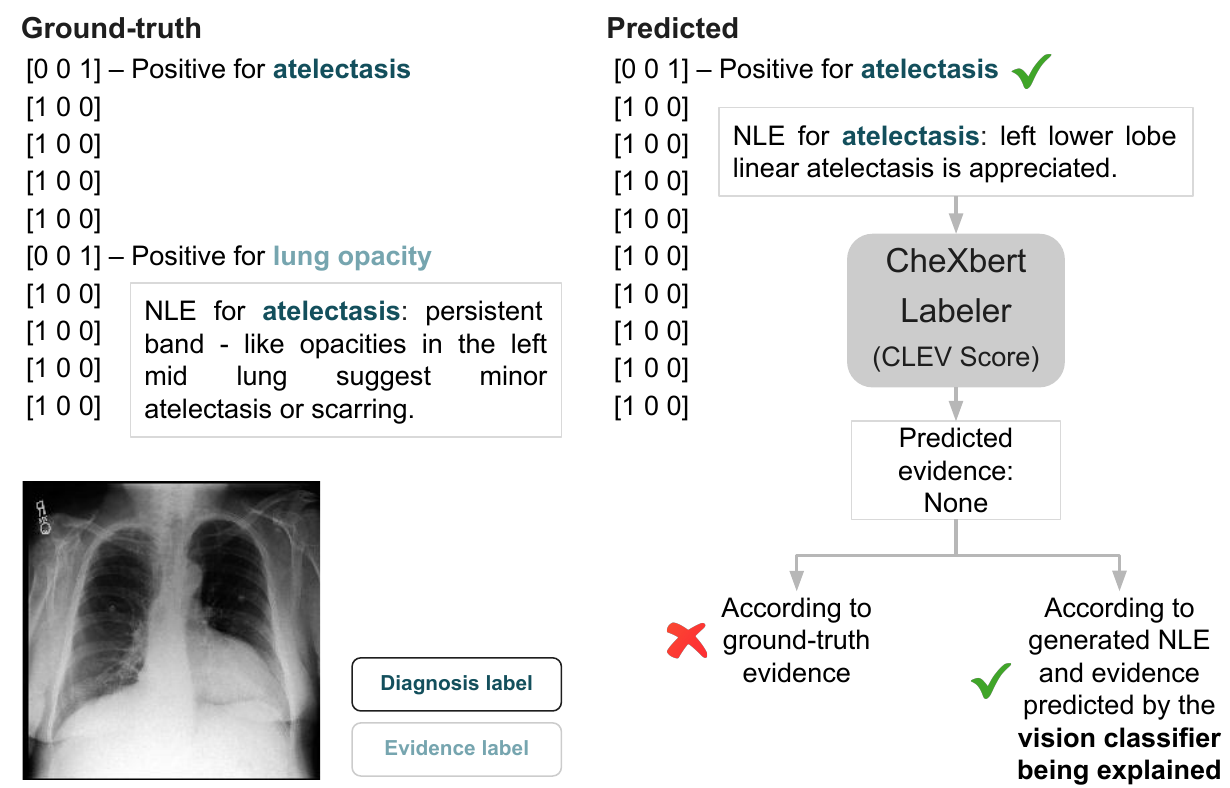}
\caption{CheXbert identified evidence is correct if it matches the evidence predicted by the model being explained (MBE) (which is also the evidence used to generate the NLE in the first place), but it is incorrect if judged against the ground-truth evidence. Since we want NLEs faithful to the MBE, the target evidence for computing the CLinical EVidence (CLEV) score should be the MBE’s predicted evidence.}
\label{fig:wrong_clev}
\end{figure}

Following the evaluation protocol of the work that originally proposed the MIMIC-NLE dataset~\cite{oana}, we also compute the CLinical EVidence (\textbf{CLEV}) score, which uses the CheXbert~\cite{chexbert} model to check if two NLEs refer to the same clinical evidence. However, we introduce two modifications: i) given the big imbalance in the evidence labels of the dataset (see Subsection~\ref{subsec:dataset}), instead of reporting the accuracy, we report the macro-averaged F1-score, and ii) we do not compare the evidence of the generated NLE with the evidence of the ground-truth NLE, because the NLE needs to be faithful to the evidence the multilabel classifier predicted, and this might differ from the ground-truth evidence. An example illustrating this can be found in Fig.~\ref{fig:wrong_clev}, where the ground-truth refers to the \textit{Lung Opacity} evidence label, but the classifier did not predict this, so the generated NLE cannot and should not mention the presence of a lung opacity. Computing the score in this way has another advantage: since we are now only dependent on the input evidence, the CLEV score can be computed for all generated NLEs, and not only for the correctly predicted ones. In fact, apart from the CXBS, which directly compares against ground-truth NLEs, all other metrics presented throughout the remainder of this work are computed for all generated NLEs.

\newpage

\subsubsection{Diversity}

Finally, we measure the diversity of the generated NLEs. Although two NLEs for the same disease and similar images should be similar, at the same time, they should not be completely equal, since the NLEs also need to be image-relevant. If they were equal, one might assume that having generated the same NLEs is due to correlations and biases in the dataset and not due to the reasoning of the MBE~\cite{nlxgpt}. To measure this, we employ the retrieval-based attack proposed in~\cite{nlxgpt} (\textbf{Retrv.}): the 10 most similar images to a query image are retrieved using any feature extraction model (we use MedImageInsight~\cite{medii}) and \emph{WeNLEX} generates one NLE for each. Then, the cosine distance between these NLEs is computed using Sentence-BERT~\cite{sbert}\footnote{\url{https://huggingface.co/sentence-transformers/all-MiniLM-L6-v2}}. The lower the distance, the lower the bias, so the better the NLEs will be. To complement this retrieval-based attack, we also compute the per-diagnosis Self-BLEU score (\textbf{S-BLEU}), i.e., the BLEU-4 score between all NLEs generated for a given diagnosis. The lower the score, the more diverse the generated NLEs.

\subsection{Results}
Quantitative results are presented in Table~\ref{tab:results}. We compare \emph{WeNLEX}, in both \textit{post-hoc} and in-model versions (models 5 and 6, respectively), with the fully supervised model of Rio-Torto \textit{et al.}~\cite{midl} retrained on our modified version of the MIMIC-NLE dataset (model 1).

We compare the effect of the proposed loss functions under the \textit{post-hoc} setting (models 2-5), since it is the only setting where we are actually comparing the same NLEs (1038 generated NLEs to be precise, of which 128 have the same diagnosis and evidence labels as their ground-truth counterparts). When comparing the two strategies for promoting plausibility, adversarial learning and MMD minimization (models 2 and 3), it can be seen that MMD minimization increases the CLEV score from 4.876 to 7.601, increases $\hat{y}|NLE$ from 71.77 to 86.89, and increases CXBS from 36.21 to 50.39. Thus, MMD minimization is clearly allowing \emph{WeNLEX} to produce NLEs closer to the ground-truth NLE distribution than the adversarial approach. Moreover, the diversity of the NLEs generated by the adversarial approach is significantly lower than that of the MMD NLEs (41.58 Self-BLEU vs 10.36), which may indicate that the WGAN-GP is collapsing to very similar NLEs for each diagnosis. For these reasons, we chose to continue all remaining experiments with the MMD loss.

Introducing the feature reconstruction loss (model 4) increases the CLEV score from 7.601 to 9.303, the percentage of prediction flips from 91.95 to 92.48, and the absolute difference in the prediction from 0.3192 to 0.3236. This increase in faithfulness comes at the cost of slightly less similarity to ground-truth NLEs (CXBS decreased from 50.39 to 47.50), but it results in a significant increase in the diversity of the NLEs (from 10.36 Self-BLEU to 2.671 and from 0.6525 to 0.5258 in the retrieval metric). Therefore, we conclude that, as hypothesised, introducing the feature reconstruction loss improves the faithfulness of the NLEs. However, both $\hat{y}|NLE$ and $\hat{y}|(img,NLE)$ decrease (from 86.89 to 77.84 and from 89.21 to 86.71, respectively). This might simply be due to the fact that the NLEs of model 4 might be more different from the text that the CheXagent model was trained on, compared to the NLEs of model 3.

Adding the NLE classification loss (model 5) further increases the CLEV score and substantially increases the $\hat{y}|NLE$ score from 77.84 to 90.46. This comes at a cost of a slight decrease in diversity (e.g., Self-BLEU increases from 2.671 to 4.037), which is to be expected since the classification loss promotes that NLEs for the same disease become similar. However, the increase in CLEV score from 9.303 to 9.401 shows that it brings benefits in terms of ensuring the correct clinical evidence.

In all cases, it can be verified that adding the NLE helps CheXagent predict the label better, as its $\hat{y}|img$ accuracy is always lower than $\hat{y}|(img, NLE)$.

In the in-model scenario (model 6), we obtain the best CLEV, $\hat{y}|NLE$, and Self-BLEU metrics, outperforming the fully supervised scenario, using only 5 ground-truth NLEs per diagnosis. It also achieves the best percentage of flipped decisions amongst the weakly supervised model variants.

Both the classifier trained on its own (models 2-5) and the classifier trained together with the explanation generation part of \emph{WeNLEX} improve upon the AUC of the classifier trained together with the fully supervised pipeline of~\cite{midl} (65.13/67.34 vs 64.08). When comparing only the \textit{post-hoc} and in-model versions of \emph{WeNLEX} (models 5 and 6), we can see that AUC improves when training the classifier together with the whole model, which shows that, unlike what has been stated many times in the literature~\cite{rudin2019stop,accvsxai}, adding interpretability does not have to mean a decrease in task performance.

\subsection{Ablations}
\subsubsection{Layers for Feature Reconstruction}
We conduct an ablation study on which classifier layer to use for the feature reconstruction loss. In the original Perceptual Loss paper~\cite{perceptual}, the authors recommend using an intermediate layer for this purpose, since higher layers do not preserve shape as well as lower layers. In our case, there does not seem to clearly exist a layer that is better than the others with regard to the several metrics we evaluate, as can be seen in Table~\ref{tab:layers}.  Using higher layers, such as the output of the classification layer or the output of the Global Average Pooling (GAP) layer, leads to less diverse NLEs (Self-BLEU scores of 5.481 and 11.38, respectively). More intermediate layers, such as the output of the 4th and 3rd dense blocks, lead to lower faithfulness scores (CLEV, $\hat{y}|NLE$, $\hat{y}|(img, NLE)$, and Flip. Lower layers like the 2nd and 1st dense blocks improve the CLEV score, while keeping the diversity higher than with the highest level layers (classifier and GAP). They also provide the highest plausibility (CXBS). Thus, we choose to use denseblock2 for all other experiments in this work, since it achieves similar metrics to denseblock1, but with a decreased computational cost (exactly half the number of features).

\begin{table}[!bth]
\centering
\caption{Ablation on the layer used for the feature reconstruction loss. ``\# Feats'' is the dimension of the flattened feature map. ``classifier'' is the output layer, ``gap'' the GAP layer, and ``denseblock'' the outputs of DenseNet layers (with ``denseblock1'' closest to the input). The underlined layer is used in all other experiments. Best results are in bold; second-best are underlined.}
\label{tab:layers}
\resizebox{\columnwidth}{!}{%
\begin{tabular}{@{}cccccccccc@{}}
\toprule
\multirow{3}[5]{*}{Layer} & \multirow{3}[5]{*}{\# Feats.} & \multicolumn{3}{c}{Faithfulness} & \multicolumn{2}{c}{Simulatability} & \multicolumn{2}{c}{Diversity \textdownarrow} & Plausibility \\
\cmidrule(lr){3-5}\cmidrule(lr){6-7}\cmidrule(lr){8-9}\cmidrule(lr){10-10}
& & & \multicolumn{2}{c}{Deletion} & \multirow{2}[3]{*}{$\hat{y}|NLE$} & \multirow{2}[3]{*}{$\hat{y}|(img,NLE)$} & \multirow{2}[3]{*}{S-BLEU} & \multirow{2}[3]{*}{Retrv.} & \multirow{2}[3]{*}{CXBS}\\
\cmidrule(lr){4-5}
& & CLEV (F1) & Flip (\%) & $\Delta p$\\
\midrule
classifier & 30 & \textbf{9.623} & 91.81 & 0.3377 & \underline{91.81} & \textbf{89.31} & 5.481 & 0.6580 & 46.96\\
gap & 1024 & 7.993 & \textbf{92.39} & 0.3279 & \textbf{92.49} & 88.34 & 11.38 & 0.6927 & 46.98\\
denseblock4 & 50176 & 8.887 & 91.62 & \textbf{0.3483} & 86.13 & 88.05  & \underline{2.424} & \underline{0.5907} & 46.80\\
denseblock3 & 200704 & 8.631 & 91.16 & 0.3344 & 86.03 & 87.57 & \textbf{1.992} & \textbf{0.5767} & 46.46\\
\underline{denseblock2} & 401408 & 9.401 & 91.48 & 0.3328 & 90.46 & \underline{88.82}  & 4.037 & 0.6286 & \underline{47.35}\\
denseblock1 & 802816 & \underline{9.463} & \underline{92.09} &	\underline{0.3385} & 89.31 & 88.63  & 3.654 & 0.6231 & \textbf{47.41}\\
\bottomrule
\end{tabular}%
}
\end{table}

\begin{table*}[!htb]
\centering
\caption{Ablation on the number of NLEs per diagnosis present in the GT NLE database. The underlined \# NLEs is used in all other experiments. Best results are in bold; second-best are underlined.}
\label{tab:nles_db}
\resizebox{0.9\textwidth}{!}{%
\begin{tabular}{@{}ccccccccc@{}}
\toprule
\multirow{3}[4]{*}{\# NLEs} & \multicolumn{3}{c}{Faithfulness} & \multicolumn{2}{c}{Simulatability} & \multicolumn{2}{c}{Diversity \textdownarrow} & Plausibility\\
\cmidrule(lr){2-4}\cmidrule(lr){5-6}\cmidrule(lr){7-8}\cmidrule(lr){9-9}
& \multirow{2}[3]{*}{CLEV (F1)} & \multicolumn{2}{c}{Deletion} & \multirow{2}[3]{*}{$\hat{y}|NLE$} & \multirow{2}[3]{*}{$\hat{y}|(img,NLE)$} & \multirow{2}[3]{*}{S-BLEU} & \multirow{2}[3]{*}{Retrv.} & \multirow{2}[3]{*}{CXBS}\\
\cmidrule(lr){3-4}
& & Flip (\%) & $\Delta p$\\
\midrule
2 & 8.663 (0.9852) & \textbf{92.76} (0.2796) & 0.3530 (0.0342) & 85.61 (4.097) & 85.64 (0.6675) & 8.544 (1.727) & 0.6852 (0.0382) & \textbf{47.40} (1.672)\\
\underline{5} & \textbf{9.615} (1.157) & 92.14 (0.2282) & 0.3539 (0.0324) & 91.78 (1.404) & \textbf{88.63} (0.4199) & 3.560 (1.551) & 0.6331 (0.0158) & 44.88 (3.371)\\
10 & 8.396 (0.1789) & 92.33 (0.4732) & \textbf{0.3749} (0.0016) & \underline{92.29} (3.424) & \underline{88.34} (0.3474) & \textbf{1.827} (0.0461) & \underline{0.6276} (0.0348) & 45.22 (1.422)\\
20 & \underline{8.672} (1.735) &  \underline{92.52} (0.4358) & \underline{0.3616} (0.0106) & \textbf{92.77} (2.569) & 88.05 (0.4199) & \underline{2.090} (0.6986) & \textbf{0.5976} (0.0400) & \underline{46.45} (0.9763)\\

\bottomrule
\end{tabular}%
}
\end{table*}

\subsubsection{Size of Ground-truth NLE Database}
We also ablate the number of clinician-annotated (i.e., ground-truth) NLEs per diagnosis label present in the NLE database. We test \emph{WeNLEX} with 2, 5, 10, and 20 NLEs per diagnosis label. Each experiment is run with 3 different random seeds, and the results (mean and standard deviations) are reported in Table~\ref{tab:nles_db}.

As expected, using only 2 NLEs per diagnosis label achieves the lowest diversity (Self-BLEU of 8.544 and retrieval score of 0.6852). It also achieves the lowest $\hat{y}|NLE$, $\hat{y}|(img, NLE)$, and $\Delta p$. Using 5, 10 or 20 NLEs does not yield significant differences, which attests to \emph{WeNLEX}'s stability and robustness without overreliance on the ground-truth NLEs. Given that using 5 NLEs achieves the highest CLEV score and a lower annotation burden than using more NLEs, all other experiments in this work are performed with 5 NLEs per diagnosis label in the database.

\subsection{Audience-Adaptable NLEs}

By not needing one NLE per image per diagnosis and simply using a few NLEs per diagnosis, not only does \emph{WeNLEX} significantly lower annotation costs while keeping the generated NLEs faithful, but it also easily allows for the adaptation of the style of the generated NLEs; one simply has to switch the NLEs in the database. This tackles another very important desirable property of NLEs: adaptability to the target audience~\cite{miller,situatednles}.

To test \emph{WeNLEX}'s ability to generate NLEs for a different target audience (i.e., not radiologists) we ask GPT4-o~\cite{gpt4o} to convert our NLE database of 5 NLEs per diagnosis label into sentences understandable by lay people, following the prompt used in Zhao \textit{et al.}~\cite{xraysimplelay}, and then use this converted database to train another version of \emph{WeNLEX}. 

\begin{figure}[!tbh]
\centerline{\includegraphics[width=0.98\columnwidth]{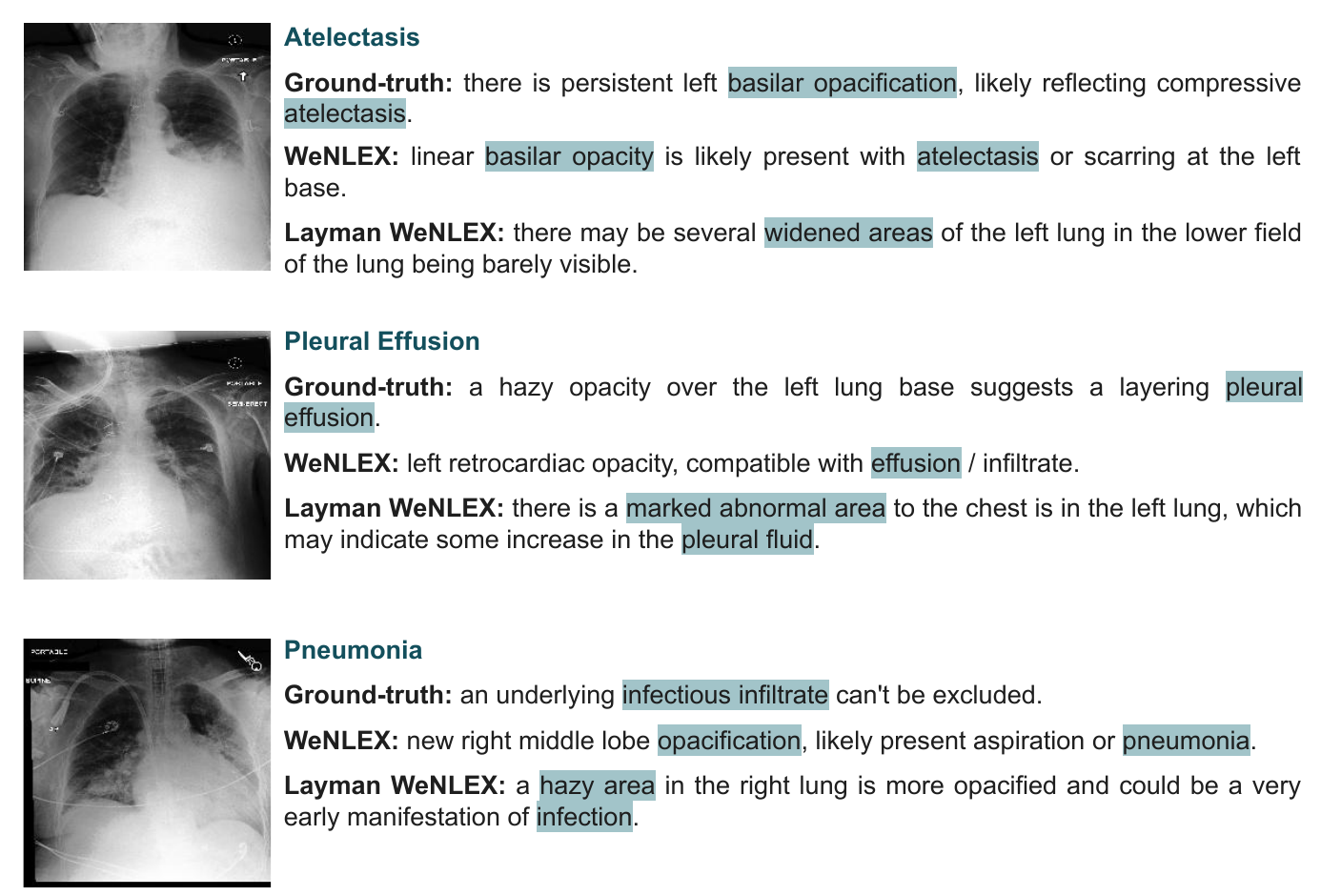}}
\caption{Qualitative examples of \emph{WeNLEX} for three different diagnoses, comparison with ground-truth NLEs, and with the layman version of \emph{WeNLEX}, in which the generated NLEs are simplified to adapt to a non-medical audience.}
\label{fig:layman}
\end{figure}

The results are presented in Table~\ref{tab:layman} and Fig.~\ref{fig:layman}. Naturally, the CLEV score decreases, since it uses the CheXbert~\cite{chexbert} labeler, which has not seen non-medical sentences during its training. Interestingly, although $\hat{y}|NLE$, $\hat{y}|(img,NLE)$, Flip and $\Delta p$ all use CheXagent~\cite{chexagent}, the last two do not seem to be as affected by the layman NLEs as the first two. This might be due to the fact that the location of the findings is given in more similar terms to the original NLEs than the findings themselves (e.g., in the first example of Fig.~\ref{fig:layman} the location cue ``at the left base'' has been switched to ``of the left lung in the lower field of the lung'', while the evidence and diagnosis, ``basilar opacity'' and ``atelectasis or scarring'', have been switched to ``widened areas'' and ``barely visible''), such that CheXagent is still able to perform the visual grounding necessary for Flip and $\Delta p$, but is no longer able to identify the diagnosis label, which is needed for $\hat{y}|NLE$ and $\hat{y}|(img, NLE)$.

As expected, the diversity is similar in both scenarios, and the plausibility decreases, since the layman NLEs are significantly different from the clinician-annotated NLEs.

Finally, we evaluate the readability of the NLEs (both those generated and those in the database). For this, we use the \texttt{textstat} Python package\footnote{\url{https://github.com/textstat/textstat}} and the \texttt{text\_standard} consensus metric, which computes the school grade level required to understand a given text, based on several tests (Flesch Reading Ease formula, Flesch-Kincaid Grade Level, Fog Scale, etc). In agreement with the proposed changes, using the simplified NLEs increases readability (the original NLEs are readable at the college level, 13.21, while the simplified NLEs are readable from 10th grade, 10.42). Thus, \emph{WeNLEX} is, in fact, able to adapt to different target audiences.

\begin{table}[ht]
\centering
\caption{Comparison of \emph{WeNLEX} trained on clinician vs. simplified layperson NLEs. The simplified version shows higher readability (lower required grade level). ``Gen.'' denotes generated NLEs, and ``DB'' the GT NLE from the medical and non-medical databases.}
\label{tab:layman}
\resizebox{\columnwidth}{!}{%
\begin{tabular}{@{}ccccccccccc@{}}
\toprule
\multirow{3}[5]{*}{Database} & \multicolumn{3}{c}{Faithfulness} & \multicolumn{2}{c}{Simulatability} & \multicolumn{2}{c}{Diversity \textdownarrow} & Plausibility & \multicolumn{2}{c}{Readability \textdownarrow} \\
\cmidrule(lr){2-4}\cmidrule(lr){5-6}\cmidrule(lr){7-8}\cmidrule(lr){9-9}\cmidrule(lr){10-11}
& \multirow{2}[3]{*}{CLEV (F1)} & \multicolumn{2}{c}{Deletion} & \multirow{2}[3]{*}{$\hat{y}|NLE$} & \multirow{2}[3]{*}{$\hat{y}|(img,NLE)$} & \multirow{2}[3]{*}{S-BLEU} & \multirow{2}[3]{*}{Retrv.} & \multirow{2}[3]{*}{CXBS} & \multirow{2}[3]{*}{Gen.} & \multirow{2}[3]{*}{DB}\\ 
\cmidrule(lr){3-4}
& & Flip (\%) & $\Delta p$\\
\midrule
Medical & 9.401 & 91.48 & 0.3328 & 90.46 & 88.82 & 4.037 & 0.6286 & 47.35 & 13.21 & 15.00\\
Layman & 4.025 & 91.43 & 0.2842 & 64.35 & 84.30 & 4.534 & 0.5295 & 39.49 & 10.42 & 10.46\\

\bottomrule
\end{tabular}%
}
\end{table}

\section{Conclusion}
We proposed \emph{WeNLEX}, a flexible weakly supervised framework for generating natural language explanations for multilabel chest X-ray classification. Through image feature reconstruction and distribution matching, its explanations are both faithful to the classifier’s reasoning and clinically plausible. \emph{WeNLEX}'s versatility allows it to work in a \textit{post-hoc} or an in-model manner. Beyond explainability, the in-model version of \emph{WeNLEX} increases AUC by 2.21\%, demonstrating that explainability and task performance can go hand in hand. \emph{WeNLEX} also enables explanations to be tailored to different audiences, from clinicians to lay users, highlighting its broad applicability in real-world medical settings. 

\bibliographystyle{IEEEtran}
\bibliography{main}

@InProceedings{gve,
    author="Hendricks, Lisa Anne
    and Akata, Zeynep
    and Rohrbach, Marcus
    and Donahue, Jeff
    and Schiele, Bernt
    and Darrell, Trevor",
    title={{Generating Visual Explanations}},
    booktitle="ECCV",
    year="2016",
    pages="3--19",
}

@InProceedings{ibpria,
    author="Rio-Torto, Isabel
    and Cardoso, Jaime S.
    and Teixeira, Lu{\'i}s F.",
    title={{From Captions to Explanations: A Multimodal Transformer-based Architecture for Natural Language Explanation Generation}},
    booktitle="Pattern Recognition and Image Analysis",
    year="2022",
    pages="54--65",
}

@article{miller,
    title = {{Explanation in artificial intelligence: Insights from the social sciences}},
    journal = {Artificial Intelligence},
    volume = {267},
    pages = {1-38},
    year = {2019},
    author = {Tim Miller},
}

@article{situatednles,
      title={{Situated Natural Language Explanations}}, 
      author={Zining Zhu and Haoming Jiang and Jingfeng Yang and Sreyashi Nag and Chao Zhang and Jie Huang and Yifan Gao and Frank Rudzicz and Bing Yin},
      year={2024},
      journal={arXiv:2308.14115},
}

@InProceedings{rexc,
  title = 	 {{Knowledge-Grounded Self-Rationalization via Extractive and Natural Language Explanations}},
  author =       {Majumder, Bodhisattwa Prasad and Camburu, Oana and Lukasiewicz, Thomas and Mcauley, Julian},
  booktitle = 	 {ICML},
  pages = 	 {14786--14801},
  year = 	 {2022},
  volume = 	 {162},
}

@InProceedings{evil,
    author    = {Kayser, Maxime and Camburu, Oana-Maria and Salewski, Leonard and Emde, Cornelius and Do, Virginie and Akata, Zeynep and Lukasiewicz, Thomas},
    title     = {{e-ViL: A Dataset and Benchmark for Natural Language Explanations in Vision-Language Tasks}},
    booktitle = {ICCV},
    year      = {2021},
    pages     = {1244-1254}
}

@article{esnlive,
    title={{e-SNLI-VE: Corrected Visual-Textual Entailment with Natural Language Explanations}},
    author={Do, Virginie and Camburu, Oana-Maria and Akata, Zeynep and Lukasiewicz, Thomas},
    journal={arXiv:2004.03744},
    year={2020}
}

@inproceedings{esnli,
    author = {Camburu, Oana-Maria and Rockt\"{a}schel, Tim and Lukasiewicz, Thomas and Blunsom, Phil},
    booktitle = {NeurIPS},
    pages = {},
    title = {{e-SNLI: Natural Language Inference with Natural Language Explanations}},
    volume = {31},
    year = {2018}
}

@inproceedings{cage,
    title = {{Explain Yourself! Leveraging Language Models for Commonsense Reasoning}},
    author = "Rajani, Nazneen Fatema  and
      McCann, Bryan  and
      Xiong, Caiming  and
      Socher, Richard",
    booktitle = {ACL},
    year = "2019",
    pages = "4932--4942"
}

@inproceedings{comve,
    title = {{{S}em{E}val-2020 Task 4: Commonsense Validation and Explanation}},
    author = "Wang, Cunxiang  and
      Liang, Shuailong  and
      Jin, Yili  and
      Wang, Yilong  and
      Zhu, Xiaodan  and
      Zhang, Yue",
    booktitle = "Workshop on Semantic Evaluation",
    year = "2020",
    pages = "307--321",
}

@inproceedings{ecqa,
    title = {{{E}xplanations for {C}ommonsense{QA}: {N}ew {D}ataset and {M}odels}},
    author = "Aggarwal, Shourya  and
      Mandowara, Divyanshu  and
      Agrawal, Vishwajeet  and
      Khandelwal, Dinesh  and
      Singla, Parag  and
      Garg, Dinesh",
    booktitle = "ACL",
    year = "2021",
    pages = "3050--3065",
}

@inproceedings{datasets,
    author = {{Wiegreffe, Sarah and Marasovi{\'c}, Ana}},
    booktitle = {NeurIPS Track on Datasets and Benchmarks},
    title = {{Teach Me to Explain: A Review of Datasets for Explainable Natural Language Processing}},
    volume = {1},
    year = {2021}
}

@InProceedings{pjx,
    author = {Park, Dong Huk and Hendricks, Lisa Anne and Akata, Zeynep and Rohrbach, Anna and Schiele, Bernt and Darrell, Trevor and Rohrbach, Marcus},
    title = {{Multimodal Explanations: Justifying Decisions and Pointing to the Evidence}},
    booktitle = {CVPR},
    year = {2018}
}

@inproceedings{fme,
    title = {{Faithful Multimodal Explanation for Visual Question Answering}},
    author = "Wu, Jialin  and
      Mooney, Raymond",
    booktitle = "ACL Workshop BlackboxNLP",
    year = "2019",
    pages = "103--112",
}

@inproceedings{rvt,
    title={{Natural Language Rationales with Full-Stack Visual Reasoning: From Pixels to Semantic Frames to Commonsense Graphs}},
    author={Marasovi{\'c}, Ana and Bhagavatula, Chandra and Park, Jae Sung and Le Bras, Ronan and Smith, Noah A and Choi, Yejin},
    booktitle={EMNLP},
    pages={2810--2829},
    year={2020}
}

@InProceedings{nlxgpt,
    author    = {Sammani, Fawaz and Mukherjee, Tanmoy and Deligiannis, Nikos},
    title     = {{NLX-GPT: A Model for Natural Language Explanations in Vision and Vision-Language Tasks}},
    booktitle = {CVPR},
    year      = {2022},
    pages     = {8322-8332}
}

@article{ofax,
      title={{Harnessing the Power of Multi-Task Pretraining for Ground-Truth Level Natural Language Explanations}}, 
      author={Björn Plüster and Jakob Ambsdorf and Lukas Braach and Jae Hee Lee and Stefan Wermter},
      year={2023},
      journal={arXiv:2212.04231},
}

@InProceedings{oana,
    author="Kayser, Maxime
    and Emde, Cornelius
    and Camburu, Oana-Maria
    and Parsons, Guy
    and Papiez, Bartlomiej
    and Lukasiewicz, Thomas",
    title={{Explaining Chest X-Ray Pathologies in Natural Language}},
    booktitle="MICCAI",
    year="2022",
    pages="701--713",
}

@article{mimic,
    title={{MIMIC-CXR, a de-identified publicly available database of chest radiographs with free-text reports}},
    author={Johnson, Alistair EW and Pollard, Tom J and Berkowitz, Seth J and Greenbaum, Nathaniel R and Lungren, Matthew P and Deng, Chih-ying and Mark, Roger G and Horng, Steven},
    journal={Scientific Data},
    volume={6},
    number={1},
    pages={317},
    year={2019},
}

@article{padchest,
    title = {{PadChest: A large chest x-ray image dataset with multi-label annotated reports}},
    journal = {Medical Image Analysis},
    volume = {66},
    pages = {101797},
    year = {2020},
    author = {Aurelia Bustos and Antonio Pertusa and Jose-Maria Salinas and Maria {de la Iglesia-Vayá}},
}

@inproceedings{interpretcxr,
    title = {{Overview of the First Shared Task on Clinical Text Generation: {RRG}24 and ``Discharge Me!''}},
    author = "Xu, Justin  and
      Chen, Zhihong  and
      Johnston, Andrew  and
      Blankemeier, Louis  and
      Varma, Maya  and
      Hom, Jason  and
      Collins, William J.  and
      Modi, Ankit  and
      Lloyd, Robert  and
      Hopkins, Benjamin  and
      Langlotz, Curtis  and
      Delbrouck, Jean-Benoit",
    booktitle = "23rd Workshop on Biomedical NLP",
    year = "2024",
    pages = "85--98",
}

@article{pubmedbert,
    author = {Gu, Yu and Tinn, Robert and Cheng, Hao and Lucas, Michael and Usuyama, Naoto and Liu, Xiaodong and Naumann, Tristan and Gao, Jianfeng and Poon, Hoifung},
    title = {{Domain-Specific Language Model Pretraining for Biomedical Natural Language Processing}},
    year = {2021},
    volume = {3},
    number = {1},
    journal = {ACM Transactions on Computing for Healthcare},
    articleno = {2},
    numpages = {23},
}

@InProceedings{densenet,
author = {Huang, Gao and Liu, Zhuang and van der Maaten, Laurens and Weinberger, Kilian Q.},
title = {{Densely Connected Convolutional Networks}},
booktitle = {CVPR},
year = {2017}
}

@inproceedings{autobot,
    title = {{Sentence Bottleneck Autoencoders from Transformer Language Models}},
    author = "Montero, Ivan  and
      Pappas, Nikolaos  and
      Smith, Noah A.",
    booktitle = "EMNLP",
    year = "2021",
    pages = "1822--1831",
}

@inproceedings{sbert,
    title = {{Sentence-BERT: Sentence Embeddings using Siamese BERT-Networks}},
    author = "Reimers, Nils and Gurevych, Iryna",
    booktitle = "EMNLP",
    year = "2019",
}

@inproceedings{chexbert,
    title = {{Combining Automatic Labelers and Expert Annotations for Accurate Radiology Report Labeling Using {BERT}}},
    author = "Smit, Akshay  and
      Jain, Saahil  and
      Rajpurkar, Pranav  and
      Pareek, Anuj  and
      Ng, Andrew  and
      Lungren, Matthew",
    booktitle = "EMNLP",
    year = "2020",
    pages = "1500--1519",
}

@inproceedings{chexagent,
    title={{CheXagent: Towards a Foundation Model for Chest X-Ray Interpretation}},
    author={Zhihong Chen and Maya Varma and Jean-Benoit Delbrouck and Magdalini Paschali and Louis Blankemeier and Dave Van Veen and Jeya Maria Jose Valanarasu and Alaa Youssef and Joseph Paul Cohen and Eduardo Pontes Reis and Emily Tsai and Andrew Johnston and Cameron Olsen and Tanishq Mathew Abraham and Sergios Gatidis and Akshay S Chaudhari and Curtis Langlotz},
    booktitle={AAAI 2024 Spring Symposium on Clinical Foundation Models},
    year={2024},
}

@InProceedings{cxrbert,
    author="Boecking, Benedikt
    and Usuyama, Naoto
    and Bannur, Shruthi
    and Castro, Daniel C.
    and Schwaighofer, Anton
    and Hyland, Stephanie
    and Wetscherek, Maria
    and Naumann, Tristan
    and Nori, Aditya
    and Alvarez-Valle, Javier
    and Poon, Hoifung
    and Oktay, Ozan",
    title={{Making the Most of Text Semantics to Improve Biomedical Vision--Language Processing}},
    booktitle="ECCV",
    year="2022",
    pages="1--21",
}

@article{chexpertplus,
    title={{CheXpert Plus: Augmenting a Large Chest X-ray Dataset with Text Radiology Reports, Patient Demographics and Additional Image Formats}},
    author={Chambon, Pierre and Delbrouck, Jean-Benoit and Sounack, Thomas and Huang, Shih-Cheng and Chen, Zhihong and Varma, Maya and Truong, Steven QH and Chuong, Chu The and Langlotz, Curtis P},
    journal={arXiv:2405.19538},
    year={2024}
}

@article{bimcv,
    title={{BIMCV COVID-19+: a large annotated dataset of RX and CT images from COVID-19 patients}},
    author={Vay{\'a}, Maria De La Iglesia and Saborit, Jose Manuel and Montell, Joaquim Angel and Pertusa, Antonio and Bustos, Aurelia and Cazorla, Miguel and Galant, Joaquin and Barber, Xavier and Orozco-Beltr{\'a}n, Domingo and Garc{\'\i}a-Garc{\'\i}a, Francisco and others},
    journal={arXiv:2006.01174},
    year={2020}
}

@article{mimiciii,
    author = {Johnson, Alistair E. W. and Pollard, Tom J. and Shen, Lu and Lehman, Li-wei H. and Feng, Mengling and Ghassemi, Mohammad and Moody, Benjamin and Szolovits, Peter and Anthony Celi, Leo and Mark, Roger G.},
    title = {{MIMIC-III, a freely accessible critical care database}},
    journal = {Scientific Data},
    volume = {3},
    number = {1},
    pages = {160035},
    year = {2016},
    type = {Journal Article}
    }

@article{pubmed, 
    title={PubMed},
    url={https://pubmed.ncbi.nlm.nih.gov/}, 
    author={{U.S. National Library of Medicine}}
}

@inproceedings{meteor,
    author = {Lavie, Alon and Agarwal, Abhaya},
    title = {{METEOR: An Automatic Metric for MT Evaluation with Improved Correlation with Human Judgments}},
    year = {2007},
    booktitle = {2nd Workshop on Statistical Machine Translation},
    pages = {228–231},
    numpages = {4},
}

@inproceedings{bertscore,
    title={{BERTScore: Evaluating Text Generation with BERT}},
    author={Tianyi Zhang and Varsha Kishore and Felix Wu and Kilian Q. Weinberger and Yoav Artzi},
    booktitle={ICLR},
    year={2020},
}

@inproceedings{papineni2002bleu,
    author = {Papineni, Kishore and Roukos, Salim and Ward, Todd and Zhu, Wei-Jing},
    title = {{BLEU: a method for automatic evaluation of machine translation}},
    year = {2002},
    booktitle = {ACL},
    pages = {311–318},
    numpages = {8},
}

@inproceedings{llama_adapter,
    title={{LLaMA-Adapter: Efficient Fine-tuning of Large Language Models with Zero-initialized Attention}},
    author={Renrui Zhang and Jiaming Han and Chris Liu and Aojun Zhou and Pan Lu and Yu Qiao and Hongsheng Li and Peng Gao},
    booktitle={ICLR},
    year={2024},
}

@INPROCEEDINGS{lossweights,
  author={Cipolla, Roberto and Gal, Yarin and Kendall, Alex},
  booktitle={CVPR}, 
  title={{Multi-task Learning Using Uncertainty to Weigh Losses for Scene Geometry and Semantics}}, 
  year={2018},
  volume={},
  number={},
  pages={7482-7491},
}

@InProceedings{midl,
  title = 	 {{Parameter-Efficient Generation of Natural Language Explanations for Chest X-ray Classification}},
  author =       {Rio-Torto, Isabel and Cardoso, Jaime S and Teixeira, Luis Filipe},
  booktitle = 	 {MIDL},
  pages = 	 {1267--1281},
  year = 	 {2024},
  volume = 	 {250},
  series = 	 {PMLR},
}

@inproceedings{wojciechowski,
    title = {{Faithful and Plausible Natural Language Explanations for Image Classification: A Pipeline Approach}},
    author = "Wojciechowski, Adam  and
      Lango, Mateusz  and
      Dusek, Ondrej",
    booktitle = "EMNLP",
    year = "2024",
    pages = "2340--2351",
}

@article{gift,
      title={{GIFT: A Framework for Global Interpretable Faithful Textual Explanations of Vision Classifiers}}, 
      author={Éloi Zablocki and Valentin Gerard and Amaia Cardiel and Eric Gaussier and Matthieu Cord and Eduardo Valle},
      year={2025},
      journal={arXiv:2411.15605}
}

@article{xraysimplelay,
      title={{X-ray Made Simple: Lay Radiology Report Generation and Robust Evaluation}}, 
      author={Kun Zhao and Chenghao Xiao and Sixing Yan and Haoteng Tang and William K. Cheung and Noura Al Moubayed and Liang Zhan and Chenghua Lin},
      year={2025},
      journal={arXiv:2406.17911},
}

@article{gpt4o,
      title={{GPT-4o System Card}}, 
      author={OpenAI},
      year={2024},
      journal={arXiv:2410.21276},
}

@inproceedings{wgan,
author = {Arjovsky, Martin and Chintala, Soumith and Bottou, L\'{e}on},
title = {{Wasserstein Generative Adversarial Networks}},
year = {2017},
booktitle = {ICML},
pages = {214–223},
numpages = {10},
}

@inproceedings{wgangp,
author = {Gulrajani, Ishaan and Ahmed, Faruk and Arjovsky, Martin and Dumoulin, Vincent and Courville, Aaron},
title = {{Improved Training of Wasserstein GANs}},
year = {2017},
booktitle = {NeurIPS},
pages = {5769–5779},
numpages = {11},
}

@InProceedings{coco,
    author="Lin, Tsung-Yi
    and Maire, Michael
    and Belongie, Serge
    and Hays, James
    and Perona, Pietro
    and Ramanan, Deva
    and Doll{\'a}r, Piotr
    and Zitnick, C. Lawrence",
    title={{Microsoft COCO: Common Objects in Context}},
    booktitle="ECCV",
    year="2014",
    pages="740--755",
}

@inproceedings{sharma2018cc3m,
    title = {{Conceptual Captions: A Cleaned, Hypernymed, Image Alt-text Dataset For Automatic Image Captioning}},
    author = "Sharma, Piyush  and
      Ding, Nan  and
      Goodman, Sebastian  and
      Soricut, Radu",
    booktitle = "ACL",
    year = "2018",
    pages = "2556--2565",
}

@INPROCEEDINGS{changpinyo2021cc12m,
  author={Changpinyo, Soravit and Sharma, Piyush and Ding, Nan and Soricut, Radu},
  booktitle={CVPR}, 
  title={{Conceptual 12M: Pushing Web-Scale Image-Text Pre-Training To Recognize Long-Tail Visual Concepts}}, 
  year={2021},
  volume={},
  number={},
  pages={3557-3567},
}

@article{medii,
  title={{MedImageInsight: An Open-Source Embedding Model for General Domain Medical Imaging}},
  author={Codella, Noel CF and Jin, Ying and Jain, Shrey and Gu, Yu and Lee, Ho Hin and Abacha, Asma Ben and Santamaria-Pang, Alberto and Guyman, Will and Sangani, Naiteek and Zhang, Sheng and others},
  journal={arXiv:2410.06542},
  year={2024}
}

@InProceedings{perceptual,
author="Johnson, Justin
and Alahi, Alexandre
and Fei-Fei, Li",
title={{Perceptual Losses for Real-Time Style Transfer and Super-Resolution}},
booktitle="ECCV",
year="2016",
pages="694--711",
}

@inproceedings{oanafaith1,
    title = {{Faithfulness Tests for Natural Language Explanations}},
    author = "Atanasova, Pepa  and
      Camburu, Oana-Maria  and
      Lioma, Christina  and
      Lukasiewicz, Thomas  and
      Simonsen, Jakob Grue  and
      Augenstein, Isabelle",
    booktitle = "ACL",
    year = "2023",
    pages = "283--294",
}

@inproceedings{oanafaith2,
    title = {{The Probabilities Also Matter: A More Faithful Metric for Faithfulness of Free-Text Explanations in Large Language Models}},
    author = "Siegel, Noah  and
      Camburu, Oana-Maria  and
      Heess, Nicolas  and
      Perez-Ortiz, Maria",
    booktitle = "ACL",
    year = "2024",
    pages = "530--546",
}

@article{faithvsplaus,
      title={{Faithfulness vs. Plausibility: On the (Un)Reliability of Explanations from Large Language Models}}, 
      author={Chirag Agarwal and Sree Harsha Tanneru and Himabindu Lakkaraju},
      year={2024},
      journal={arXiv:2402.04614},
}

@inproceedings{las,
    title = {{Leakage-Adjusted Simulatability: Can Models Generate Non-Trivial Explanations of Their Behavior in Natural Language?}},
    author = "Hase, Peter  and
      Zhang, Shiyue  and
      Xie, Harry  and
      Bansal, Mohit",
    booktitle = "EMNLP",
    year = "2020",
    pages = "4351--4367",
}

@inproceedings{wiegreffe21,
    title = {{Measuring Association Between Labels and Free-Text Rationales}},
    author = {{Wiegreffe, Sarah  and
      Marasovi{\'c}, Ana  and
      Smith, Noah A.}},
    booktitle = "EMNLP",
    year = "2021",
    pages = "10266--10284",
}

@inproceedings{PetsiukDS18,
  author       = {Vitali Petsiuk and
                  Abir Das and
                  Kate Saenko},
  title        = {{{RISE:} Randomized Input Sampling for Explanation of Black-box Models}},
  booktitle    = {BMVC},
  pages        = {151},
  year         = {2018},
}

@article{deletion,
    title = {{Methods for interpreting and understanding deep neural networks}},
    journal = {Digital Signal Processing},
    volume = {73},
    pages = {1-15},
    year = {2018},
    author = {Grégoire Montavon and Wojciech Samek and Klaus-Robert Müller},
}

@article{accvsxai,
      title={{Demystifying the Accuracy-Interpretability Trade-Off: A Case Study of Inferring Ratings from Reviews}}, 
      author={Pranjal Atrey and Michael P. Brundage and Min Wu and Sanghamitra Dutta},
      year={2025},
      journal={arXiv:2503.07914}, 
}

@article{rudin2019stop,
  title={{Stop explaining black box machine learning models for high stakes decisions and use interpretable models instead}},
  author={Rudin, Cynthia},
  journal={Nature Machine Intelligence},
  volume={1},
  number={5},
  pages={206--215},
  year={2019},
}

@inproceedings{jacovi,
    title={{Towards Faithfully Interpretable {NLP} Systems: How Should We Define and Evaluate Faithfulness?}},
    author = "Jacovi, Alon  and
      Goldberg, Yoav",
    booktitle = "ACL",
    year = "2020",
    pages = "4198--4205",
}

@article{beenkim2017,
      title={{Towards A Rigorous Science of Interpretable Machine Learning}}, 
      author={Finale Doshi-Velez and Been Kim},
      year={2017},
      journal={arXiv:1702.08608}, 
}

@inproceedings{zsnles,
    title={{Zero-Shot Natural Language Explanations}},
    author={Fawaz Sammani and Nikos Deligiannis},
    booktitle={ICLR},
    year={2025},
}

@InProceedings{unsupcapt,
    author = {Feng, Yang and Ma, Lin and Liu, Wei and Luo, Jiebo},
    title = {{Unsupervised Image Captioning}},
    booktitle = {CVPR},
    year = {2019}
}

@article{kgllava,
    title={{LLaVA Needs More Knowledge: Retrieval Augmented Natural Language Generation with Knowledge Graph for Explaining Thoracic Pathologies}}, 
    volume={39}, 
    number={3}, 
    journal={AAAI Conference on Artificial Intelligence}, 
    author={Hamza, Ameer and Abdullah, Abdullah and Ahn, Yong Hyun and Lee, Sungyoung and Kim, Seong Tae}, 
    year={2025},
    pages={3311--3319}
}

@InProceedings{llava15,
    author    = {Liu, Haotian and Li, Chunyuan and Li, Yuheng and Lee, Yong Jae},
    title     = {Improved Baselines with Visual Instruction Tuning},
    booktitle = {CVPR},
    year      = {2024},
    pages     = {26296-26306}
}

@INPROCEEDINGS{lora,
  author={Yu, Yu and Yang, Chao-Han Huck and Kolehmainen, Jari and Shivakumar, Prashanth G. and Gu, Yile and Ren, Sungho Ryu Roger and Luo, Qi and Gourav, Aditya and Chen, I-Fan and Liu, Yi-Chieh and Dinh, Tuan and Filimonov, Ankur Gandhe Denis and Ghosh, Shalini and Stolcke, Andreas and Rastow, Ariya and Bulyko, Ivan},
  booktitle={IEEE Automatic Speech Recognition and Understanding Workshop}, 
  title={{Low-Rank Adaptation of Large Language Model Rescoring for Parameter-Efficient Speech Recognition}}, 
  year={2023},
  pages={1-8},
}

@inproceedings{radgraph,
 author = {Jain, Saahil and Agrawal, Ashwin and Saporta, Adriel and Truong, Steven and Duong, Du Nguyen Duong Nguyen and Bui, Tan and Chambon, Pierre and Zhang, Yuhao and Lungren, Matthew and Ng, Andrew and Langlotz, Curtis and Rajpurkar, Pranav and Rajpurkar, Pranav},
 booktitle = {NeurIPS Track on Datasets and Benchmarks},
 title = {{RadGraph: Extracting Clinical Entities and Relations from Radiology Reports}},
 volume = {1},
 year = {2021}
}

@INPROCEEDINGS{uninlx,
  author={Sammani, Fawaz and Deligiannis, Nikos},
  booktitle={ICCVW}, 
  title={{Uni-NLX: Unifying Textual Explanations for Vision and Vision-Language Tasks}}, 
  year={2023},
  volume={},
  number={},
  pages={4636-4641},
}

@article{Vayena,
    author = {Vayena, Effy AND Blasimme, Alessandro AND Cohen, I. Glenn},
    journal = {PLOS Medicine},
    title = {{Machine learning in medicine: Addressing ethical challenges}},
    year = {2018},
    volume = {15},
    pages = {1-4},
}

@article{gale,
  title={{Producing radiologist-quality reports for interpretable artificial intelligence}},
  author={Gale, William and Oakden-Rayner, Luke and Carneiro, Gustavo and Bradley, Andrew P and Palmer, Lyle J},
  journal={arXiv:1806.00340},
  year={2018}
}

\end{document}